\documentclass[lettersize,journal]{IEEEtran}
\usepackage{amsmath,amsfonts}
\usepackage{algorithmic}
\usepackage{algorithm}
\usepackage{array}
\usepackage[caption=false,font=normalsize,labelfont=sf,textfont=sf]{subfig}
\usepackage{textcomp}
\usepackage{stfloats}
\usepackage{url}
\usepackage{verbatim}
\usepackage{graphicx}
\usepackage{cite}
\usepackage{multirow}
\usepackage{color}
\newcommand{\todo}[1]{\textcolor{blue}{#1}}

\begin{document}

\title{Evolutionary Computation in the Era of \\Large Language Model: Survey and Roadmap}

\author{Xingyu Wu, Sheng-hao Wu*,~\IEEEmembership{Member, IEEE}, Jibin Wu*,\\  Liang Feng*,~\IEEEmembership{Senior Member, IEEE}, Kay Chen Tan,~\IEEEmembership{Fellow, IEEE}
\thanks{Xingyu Wu, Sheng-Hao Wu, Jibin Wu, and Kay Chen Tan are with the Department of Computing, The Hong Kong Polytechnic University, Hong Kong SAR 999077, China (email: xingy.wu@polyu.edu.hk, shenghao.wu@polyu.edu.hk, jibin.wu@polyu.edu.hk, kaychen.tan@polyu.edu.hk).}
\thanks{Liang Feng is with the College of Computer Science, Chongqing University, Chongqing 400044, China (e-mail: liangf@cqu.edu.cn).}
\thanks{*Corresponding author: Sheng-hao Wu, Jibin Wu, and Liang Feng.}
\thanks{Manuscript received January 15, 2024.}
}



\maketitle

\begin{abstract}
Large language models (LLMs) have not only revolutionized natural language processing but also extended their prowess to various domains, marking a significant stride towards artificial general intelligence. The interplay between LLMs and evolutionary algorithms (EAs), despite differing in objectives and methodologies, share a common pursuit of applicability in complex problems. Meanwhile, EA can provide an optimization framework for LLM's further enhancement under black-box settings, empowering LLM with flexible global search capacities. On the other hand, the abundant domain knowledge inherent in LLMs could enable EA to conduct more intelligent searches. Furthermore, the text processing and generative capabilities of LLMs would aid in deploying EAs across a wide range of tasks. Based on these complementary advantages, this paper provides a thorough review and a forward-looking roadmap, categorizing the reciprocal inspiration into two main avenues: LLM-enhanced EA and EA-enhanced LLM. Some integrated synergy methods are further introduced to exemplify the complementarity between LLMs and EAs in diverse scenarios, including code generation, software engineering, neural architecture search, and various generation tasks. As the first comprehensive review focused on the EA research in the era of LLMs, this paper provides a foundational stepping stone for understanding the collaborative potential of LLMs and EAs. The identified challenges and future directions offer guidance for researchers and practitioners to unlock the full potential of this innovative collaboration in propelling advancements in optimization and artificial intelligence. We have created a GitHub repository to index the relevant papers: \url{https://github.com/wuxingyu-ai/LLM4EC}.
\end{abstract}


\section{Introduction}

In recent years, large language models (LLMs\footnote{This paper views the LLM as the Transformer-based language model with a large number of parameters, pretrained on massive datasets using self/semi-supervised learning techniques.}) \cite{devlin2018bert,pan2024unifying,raffel2020exploring} have achieved notable research breakthroughs, showcasing remarkable performance in the field of natural language processing \cite{chowdhery2023palm}. As the scale of these models expands, LLMs showcase not only excellence in language-related tasks but also reveal expansive potential applications in diverse domains \cite{wei2022emergent}. This includes a spectrum of optimization and generation tasks, representing a pivotal milestone in the evolution of artificial general intelligence. The advancement of LLMs has also catalyzed progress in technologies and methodologies across various research field \cite{wu2023survey,chen2023large,zhou2024causalbench}. Notably, this impact extends to evolutionary computation, offering both new opportunities and challenges. The primary goal of this review is to explore the dynamic interplay and synergies between LLMs and evolutionary algorithms (EAs), with the intention of establishing a complementary relationship between the two within the contemporary era of LLMs. 



The LLM and EA, despite substantial disparities in objectives and principles, share a common pursuit of applicability in various scenarios, which are different from most models that aimed for high performance in specific domain problems. LLM achieves a unified approach across diverse tasks by learning from extensive data \cite{min2023recent}, while EA, as a general-purpose solver, has lower reliance on problem characteristics and information compared to traditional mathematical optimization methods \cite{tian2022comparative}, enabling it to solve a wider range of problems with different characteristics. Therefore, in terms of application scenarios, both EAs and LLMs demonstrate unique advantages in addressing complex problems with vast search spaces and uncertain environments \cite{zhan2022survey,wei2022emergent}. This similarity suggests potential complementarity and mutual inspiration between LLM and EA when dealing with large-scale and complex problems. 

Although LLM has achieved success in various applications, it has still faced criticism attributable to its black-box nature and inflexible searching. Due to the intricate LLM architecture, the specific details of the internal decision-making, reasoning, and generation processes are either uninterpretable or invisible for most users \cite{ray2023chatgpt}, especially in the case where commercially viable LLMs (such as GPT-4 \cite{openai2023gpt4}) typically keep their model parameters private. Exactly, as a classic black-box optimization technique \cite{povsik2012comparison}, EAs hold potential for further enhancement within the black-box framework of LLM, such as prompt optimization \cite{guo2023connecting} or neural architecture search (NAS) \cite{gao2022autobert}. Another limitation of LLM is its finite search capability, as the search process is typically conducted in a one-shot manner without iterative progressive optimization. Moreover, the search capability of LLMs is constrained by prompts and training data, leading to a tendency to generate content that aligns with learned patterns and prompt information \cite{yang2023harnessing}, thereby limiting global exploration of the entire search space. EA's search superiority can mitigate this limitation in LLM. Firstly, EA is an iterative optimization method that can continuously evolve and improve the solutions generated by LLM, thus enhancing result quality. Additionally, EA can achieve more flexible search through well-designed searching space and evolutionary operator \cite{deng2023large,romera2023mathematical}. The search capacity of EA proves particularly advantageous for complex tasks that require adequate optimization. This is highly relevant for the case of LLMs, which often require extensive tuning of hyperparameters and prompts to achieve peak performance~\cite{petke2017genetic,MarioGPT2023illuminating}.

\begin{figure}[t]
\begin{center}
\includegraphics[width=0.45\textwidth]{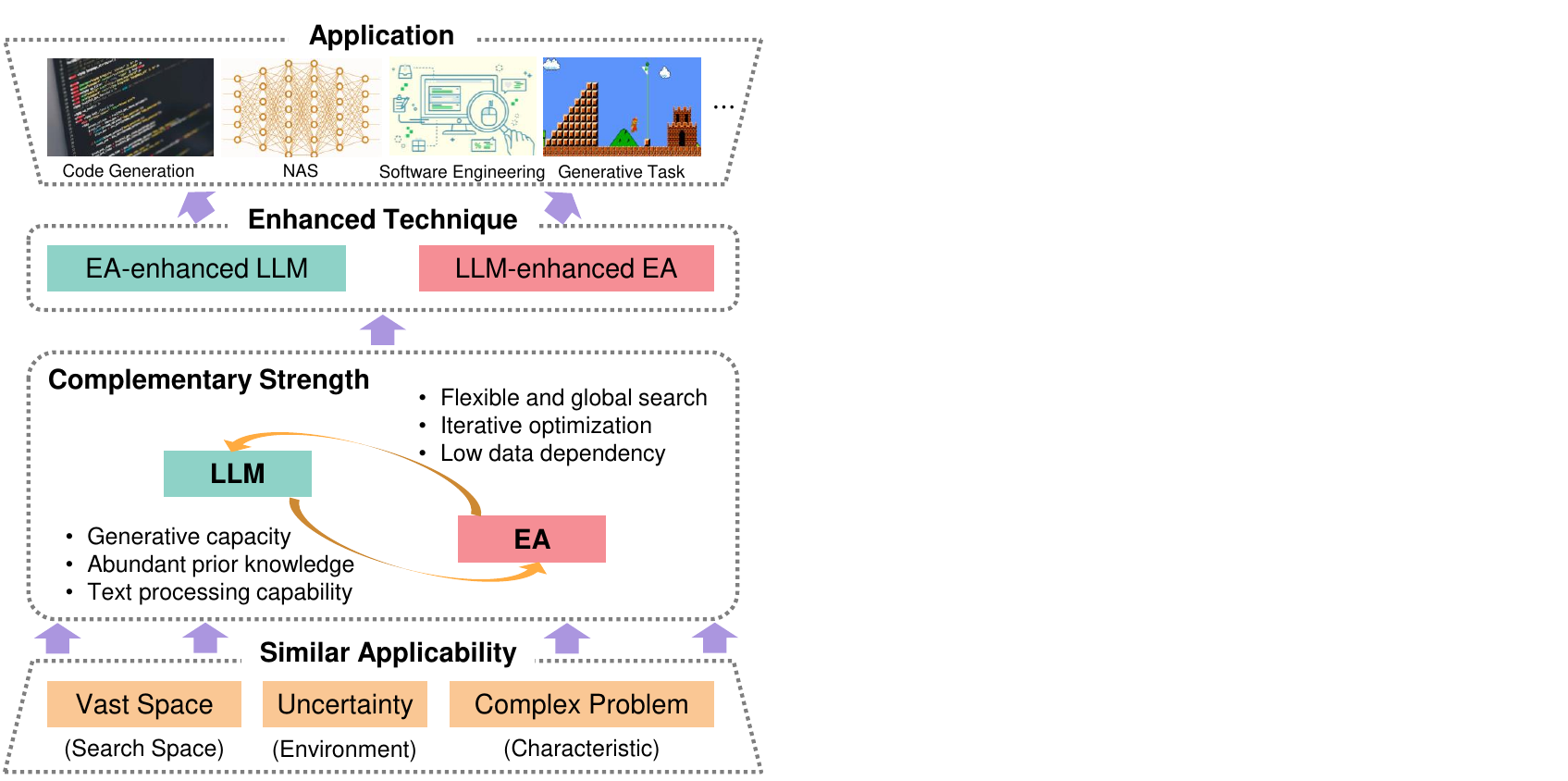}
\end{center}
\caption{A general framework for integrated research of LLMs and EAs.}
\label{Framework}
\end{figure}

On the other hand, LLMs demonstrate superiority in general knowledge, text understanding, and generative capacity. These strengths can compensate for certain limitations encountered when applying EA independently. Firstly, taking benefit from pretraining on extensive text datasets, LLMs contain a wealth of knowledge across various domains, which can provide effective guidance during the EA search process. Although EA has powerful global search capabilities, it often requires more search steps to achieve desirable results due to the lack of task-related knowledge \cite{liu2023largeb}, particularly in situations with a challenging search space or encompass a diversified population \cite{woldesenbet2009dynamic}. Research has shown that LLM can offer valuable information at the early search stage, aiding EA in faster converging to promising solutions \cite{liu2023largeb}. Additionally, another inherent strength of LLMs lies in their remarkable text understanding and generation capability. Since EA is typically designed for numerical problems and require additional encoding or preprocessing steps to adapt to text-related tasks \cite{araujo2007evolutionary}, integrating LLM with EA obviously facilitates more convenient utilization of EA's algorithmic principles in tasks involving text generation and processing \cite{xiao2023enhancing,MarioGPT2023illuminating}. Moreover, under the evolutionary framework, LLMs hold significant promise for a wide range of tasks that involve content generation, like prompt generation \cite{guo2023connecting} and algorithm generation \cite{lehman2023evolution}.

\begin{figure}[t]
\begin{center}
\includegraphics[width=0.48\textwidth]{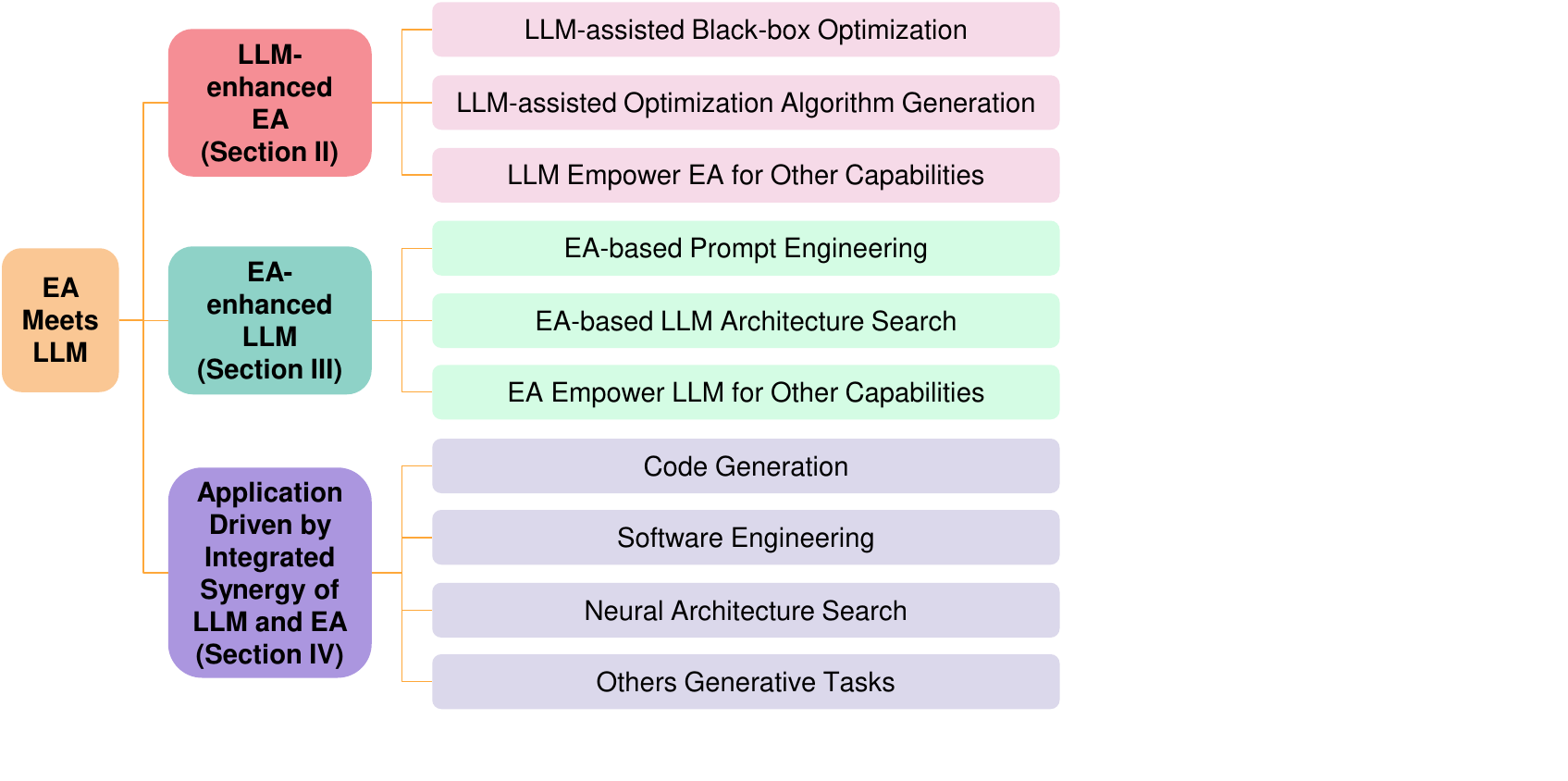}
\end{center}
\caption{Categorization of research works surveyed in this paper.}
\label{Category}
\end{figure}

Given the above complementary advantages, the potential collaboration between EA and LLM has gained increasing attention from researchers and practitioners recently. Current research primarily focuses on three aspects to enhance mutual performance and jointly drive application developments, as shown in Fig. \ref{Category}:
\begin{enumerate}
\item \emph{LLM-enhanced Evolutionary Optimization:} LLMs can serve not only as evolution operators, leveraging rich domain knowledge to accelerate the search process, but can also utilize their code generation capabilities to enhance EAs at the algorithmic level.
\item \emph{EA-enhanced LLM:} The black-box optimization advantage of EAs can aid LLMs in prompt engineering. This, in turn, enhances the outputs of LLMs with improved prompts. Additionally, the search capability of EAs can optimize the neural architecture of LLMs, resulting in versatile and lightweight LLMs.
\item \emph{Applications Driven by Integrated Synergy of LLM and EA:} The collaborative synergy between LLM and EA has revolutionized numerous conventional application scenarios, including NAS, code generation, software engineering, and text generation.
\end{enumerate}

In this paper, we present a comprehensive review and a forward-looking roadmap for the cross-disciplinary study of LLMs and EAs. We provide a detailed categorization of rapidly evolving areas, conduct a thorough review, and identify emerging directions. Our primary contributions are summarized as follows:
\begin{enumerate}
\item We present a systematic survey of the current state of cross-disciplinary research between LLMs and EAs. Through the analysis of existing methods, we comprehensively review related research progress and applications. To the best of our knowledge, this paper is the first comprehensive and up-to-date review specifically focused on the EA research in the era of LLMs.
\item We propose a comprehensive taxonomy that classifies the current cross-disciplinary research on LLMs and EAs into three distinct categories. This systematic categorization allows for a clear and organized overview of this emerging field, enabling existing methods to be appropriately placed within their respective categories.
\item Through critical analysis on the strengths and weaknesses of the existing methods, we identify several key challenges in the cross-disciplinary research of LLMs and EAs. These findings serve as a source of inspiration for future investigations in this promising area.
\end{enumerate}

The remainder of this paper is organized as follows. Sections II and III delve into the synergistic research on the complementary strengths of LLM and EA, specifically examining how LLM enhances EA and how EA enhances LLM. In Section IV, a comprehensive review is provided on the application domains that have been significantly influenced by the combined advancements of LLM and EA. Based on the extensive survey of existing research, Section V highlights several key areas that warrant future investigation and provides a roadmap for further exploration. Finally, Section VI concludes this paper. 

\section{LLM-enhanced EA}

LLMs hold immense promise for solving optimization problems. With their robust natural language understanding and generation capabilities, LLMs can effectively handle complex problem descriptions and constraints. Currently, there are two primary approaches for leveraging LLMs to assist in solving optimization problems: (1) The first approach uses LLMs as black-box search operators for optimization problems. This approach leverages the LLM's generation ability to create novel solutions. A summary of this approach can be found in Table \ref{tab:LLM4EA_table1}. (2) The second approach leverages the representation power and generation abilities of LLMs to generate novel optimization algorithms for solving specific problems, as shown in Table \ref{tab:LLM4EA_table2}.

\subsection{LLM-assisted Black-box Optimization: As Search Operator}


The potential of LLMs in solving optimization problems has been extensively demonstrated by numerous studies, including two empirical study for comprehensive evaluations \cite{guo2023towards,huang2024exploring}. In particular, LLMs as search operators in black-box optimization have emerged as a prominent research area. Several studies have validated its strong optimization abilities in solving small-sized problems. Specifically, these studies utilize LLMs to generate the next generation of solutions for both single-objective and multi-objective optimizations.

\begin{table}[t]
\centering
\caption{Summary of LLM-assisted Black-box Optimization}
\label{tab:LLM4EA_table1}
\begin{tabular}{|c|c|c|}
\hline
\textbf{Type}                              & \textbf{Method}                                                & \textbf{Ref.}                                    \\ \hline
                        & N/A                                                   & \cite{guo2023towards}       \\
\multirow{-2}{*}{Evaluation}  & N/A        & \cite{huang2024exploring}        \\
\hline
                                  & OPRO                     & \cite{yang2023large}        \\ 
                                  &  LMX & \cite{meyerson2023language} \\ 
                                  & LMEA         & \cite{liu2023largeb}        \\
                                  & EvoLLM         & \cite{lange2024large}        \\
                                  & Huang \emph{et al.}         & \cite{huang2024multimodal}        \\
                                  & LEO         & \cite{brahmachary2024large}        \\
                                  & Application for soft robots         & \cite{zhang2024cuda}        \\
\multirow{-7}{*}{Sing-objective}  & Application for visual representation         & \cite{chiquier2024evolving}        \\ \hline
                                  & Decomposition-based MOEA                    & \cite{liu2023largea}        \\ 
                                  & QDAIF         & \cite{bradley2023quality}   \\
                                  & In-context QD         & \cite{lim2024large}   \\
\multirow{-4}{*}{Multi-objective} & CMOEA-LLM         & \cite{wang2024large}   \\ \hline
\end{tabular}%
\end{table}

\subsubsection{Single-objective optimization}

\todo{Yang \emph{et al.} first discovered that LLMs have the ability to progressively improve solutions in optimization tasks when provided with the problem and past trajectory in natural language \cite{yang2023large}. They proposed Optimization by PROmpting (OPRO) to leverage LLMs as optimizers. OPRO is primarily applicable in the absence of gradients, where optimization problems are described in natural language and serve as meta-prompts. In each iteration, LLMs use the previously generated solutions and their values as prompts to generate new solutions, which are then evaluated and added to the prompt for the next iteration. While this study does not explicitly define the `crossover' and `mutation' operators in evolutionary optimization, the role of LLMs in black-box optimization, particularly OPRO's emphasis on utilizing optimization trajectories to help LLMs identify patterns of high-quality solutions, can be applied in evolutionary optimization. For instance, a similar study in evolutionary optimization is Language Model Crossover (LMX) \cite{meyerson2023language}, which employs LLMs to generate new offspring solutions from parent solutions represented as text. LMX acts as an evolutionary variation operator by prompting the LLM with concatenated parents and interpreting its output as offspring. LMX exhibits properties such as heritability of traits from parents to offspring and universality in representing any genetic operator. The authors demonstrate the performance of LMX in various scenarios.}

\todo{Similar to the aforementioned research, some researchers have made improvements to the evolutionary optimization process based on LLM from different aspects. Liu \emph{et al.} propose the LLM-driven EA (LMEA) \cite{liu2023largeb}, which not only uses LLM to perform crossover and mutation operations, but also constructs a prompt in each generation to guide the LLM in selecting parent solutions from the current population. Lange \emph{et al.} propose EvoLLM and focus on prompt design to transform LLMs into evolution strategies for black-box optimization \cite{lange2024large}. Huang \emph{et al.} further leveraged GPT-4's multi-modal capabilities \cite{huang2024multimodal}. In addition to textual prompts, they additionally provide the LLM with visual prompts representing the layout of the nodes in capacitated vehicle routing problem, to further boost performance. Brahmachary \emph{et al.} propose the Language-model-based Evolutionary Optimizer (LEO) for population-based EAs, which focuses on the balance between exploration and exploitation \cite{brahmachary2024large}. LEO divides the solution set into exploration and exploitation pools. It utilizes LLMs to generate new solutions for the two pools separately based on different prompts. Then, LEO uses an elitism selection strategy to guide the evolutionary direction, which imports the solutions with the minimum objective function values from the exploration pool into the exploitation pool, while removing solutions with the maximum objective function values from the exploitation pool. In addition, LLM-guided EAs have demonstrated practicality in various real-world applications, such as the co-design of soft robot's morphology and control \cite{zhang2024cuda}, as well as the interpretable and discriminative attribute representation for visual classification \cite{chiquier2024evolving}.}

\subsubsection{Multi-objective optimization}

In the field of multiobjective EAs (MOEAs), Liu \emph{et al.} utilize a decomposition-based MOEA framework and employ LLMs as black-box search operators to generate new offspring individuals for each subproblem \cite{liu2023largea}, through prompt engineering and in-context learning. Moreover, they further design an explicit white-box linear operator based on interpreting the behavior of the LLM to approximate its results, and validate its generalization in experiments. Another example of applying LLMs to multi-objective evolutionary optimization is Quality-Diversity (QD) search, namely QD through AI Feedback (QDAIF) algorithm \cite{bradley2023quality}. It uses LLMs to evaluate the quality and diversity of generated solutions, rather than relying on manually designed metrics, allowing it to be applied to more complex qualitative domains. The EA is responsible for maintaining the solution library, and replacing newly generated higher quality and more diverse solutions into the relevant positions in the library based on the evaluation of the LLM, realizing an iterative optimization search process. Another similar application of LLMs in QD problems is In-context QD \cite{lim2024large}.

Unlike previous methods, Wang \emph{et al.} utilized the fine-tuned LLM as an evolutionary operator to generate $10\%$ of offspring and accelerate the convergence rate of the population \cite{wang2024large}. Specifically, the training samples provided to the LLM contained feasible and infeasible solutions with different qualities, along with the decision variable values, objective function values, and constraint violation degrees of each solution. Additionally, a specialized prompt language was designed to clarify the LLM's task of producing new solutions. The prompt language emphasized that new solutions should simultaneously consider objective function optimization and constraint satisfaction. According to the prompt language, the LLM learned how to select two solutions from the input samples and generate completely new offspring based on them through recombination or other operations. Through iterative training, the LLM continuously optimized its ability to produce high-quality solutions and more efficiently solve constrained multi-objective optimization problems.


Existing research has shown that LLMs have potential for small-scale optimization problem. Compared with traditional EAs, LLM can understand the optimization problems and the expected properties of solutions using natural language, which is more direct and simple than formally defining problems and implementing operators through programming. It also avoids the need for additional training and can generalize to different problems. Moreover, the rich prior knowledge of LLM can realize operators that are difficult to design manually in algorithm design, providing stronger exploration ability for the algorithm, which may even surpass artificially designed operators in this respect. In addition to optimizing results, the utilization of evolution operators based on LLMs has shown significant benefits in terms of efficiency. \todo{Several studies \cite{meyerson2023language,liu2023largeb,lange2024large,brahmachary2024large} have demonstrated that LLM-guided evolution outperforms random optimization in terms of efficiency. Notably, research conducted by \cite{meyerson2023language,brahmachary2024large} reveals that LLM-guided evolution operators achieve search efficiency comparable to manually designed genetic operators. Furthermore, as the scale of the LLM increases, the performance of LLM-guided evolution operators improves, and their evolutionary efficiency advances alongside the progress of the LLM itself. Additionally, \cite{liu2023largeb} highlights that controlling the temperature of the LLM to balance exploration and exploitation can further enhance the search efficiency of LLM-guided evolution. Moreover, \cite{lange2024large} showcases the potential advantages of LLM-based evolution operators in terms of initial convergence speed, and efficiency can be further enhanced by incorporating optimization hints.} LLM also brings EA advantages such as flexible input and output scales, good memory of the optimization process, and better zero-shot learning effects. However, applying LLMs to practical complex optimization problems poses major challenges \cite{huang2024exploring}. Problems with high dimensions, constraints, and precision require interactions that exceed LLMs' context abilities. Additionally, current evaluations focus narrowly and consider limited factors, insufficient to demonstrate LLMs' full optimization capabilities. Overall, while initial studies are promising, significant barriers remain for applying LLMs to real-world complex optimization problems.

\subsection{LLM-assisted Optimization Algorithm Generation}
\label{llmgenerateea}

In addition to using LLMs as evolutionary operators as mentioned earlier, some studies have leveraged LLMs for algorithmic optimization. By automatically generating optimization algorithms, LLM provides powerful support for solving optimization problems. At the beginning, the code generation methods were primarily \emph{single-round}, relying on additional debug steps to optimize the code. After the emergence of \cite{lehman2023evolution}, researchers started to use EA to achieve \emph{iterative generation}.


\begin{table}[t]
\centering
\caption{Summary of LLM-assisted Optimization Algorithm Generation}
\label{tab:LLM4EA_table2}
\resizebox{0.48\textwidth}{!}{
\begin{tabular}{|c|c|c|}
\hline
\textbf{Method} & \textbf{Generated Algorithm or Target Problem} & \textbf{Ref.} \\ \hline
Pluhacek \emph{et al.} & Hybrid swarm intelligence optimization algorithm & \cite{pluhacek2023leveraging} \\ 
OptiMUS & Mixed-integer linear programming problem & \cite{ahmaditeshnizi2023optimus} \\ 
ZSO & Zoological search optimization algorithm & \cite{zhong2024leveraging} \\
AEL & Heuristic algorithm & \cite{liu2023algorithm} \\ 
ReEvo & Heuristic algorithm & \cite{ye2024reevo} \\
LLM\_GP & Genetic Programming & \cite{hemberg2024evolving} \\
SR-EAD & Evolutionary strategy or evolution transformer & \cite{lange2024evolution} \\
EvolCAF & Cost-aware Bayesian optimization & \cite{yao2024evolve} \\
Kramer & Evolution Strategies & \cite{kramer2024large} \\
\hline
\end{tabular}}
\end{table}

\subsubsection{Single-round Generation} Pluhacek \emph{et al.} used an LLM to generate hybrid swarm intelligence optimization algorithms \cite{pluhacek2023leveraging}. Specifically, the LLM first selects suitable candidate algorithms for the given problem and analyzes the components of each algorithm, then automatically designs the hybridization approach while providing reasoning and code implementation. Later, OptiMUS further develops the role of LLMs specifically for mixed-integer linear programming (MILP) problems \cite{ahmaditeshnizi2023optimus}, enabling automated optimization problem modeling and solving. It leverages LLMs to achieve automation throughout the problem-modeling and solving phases, including mathematical modeling, algorithm selection, code writing, debugging, and validity checking. Another recent study focused on the animal-inspired metaheuristic domain \cite{zhong2024leveraging}.


\subsubsection{Iterative Generaiton} In the nascent stages of LLM development, Lehman \emph{et al.} pioneered the integration of EA into the field of algorithm generation \cite{lehman2023evolution}\footnote{The method presented in \cite{lehman2023evolution} does not concentrate exclusively on the automated generation of optimization algorithms, but applicable across a wide range of domains. Given that this subsection is specifically focused on the optimization algorithm generation, the discussion of other more general methods for algorithm/code generation, including the techniques in \cite{lehman2023evolution}, will be detailed in Section \ref{code_generation}.}. They leveraged the iterative optimization framework of EA to systematically refine algorithms, employing LLMs as the evolutionary operators that directly manipulate code text. This innovative approach paved the way for a surge of subsequent studies \cite{ma2023eureka,chen2023seed,romera2023mathematical,fu2023gpt4aigchip}, which have since expanded the application of this framework to generate algorithms and code across a diverse range of domains, including the optimization domain. Following this framework, Liu \emph{et al.} proposed Algorithm Evolution using LLM (AEL) \cite{liu2023algorithm} for automated heuristic algorithm design, which treats the heuristic idea/thought as an independent evolutionary object, and optimizes it collaboratively with the code. Reflective Evolution (ReEvo) presented by Ye \emph{et al.} improves this framework by introducing short-term and long-term reflection mechanisms to better use previously generated algorithms \cite{ye2024reevo}. Other methods following this iterative manner have also emerged in different scenarios, e.g., the genetic programming (GP) \cite{hemberg2024evolving}, acquisition functions in cost-aware Bayesian optimization \cite{yao2024evolve}, and evolution strategies \cite{kramer2024large}. 

\subsubsection{EA Representation Learning-based Generation} Beside the aforementioned methods, Lange \emph{et al.} propose a EA generation method based on the algorithm representation, called Self-Referential EA Distillation (SR-EAD) \cite{lange2024evolution}, to implement EA generation based on their own trained Evolution Transformer, a model characterizing various families of evolutionary strategies. Specifically, the model processes the population members, fitnesses, and search distribution statistics of each generation to output the update of the search distribution for the next generation. This model is trained with EA Distillation to clone existing black-box optimization algorithms, such as Covariance Matrix Adaptation Evolutionary Strategy (CMA-ES) \cite{hansen2003reducing}. After training, the model can perform in-context evolutionary optimization on new tasks, demonstrating that it has learned general optimization principles. The parameters of the well-trained Evolution Transformer are randomly perturbed to generate multiple mutated model instances. These mutated model instances each run on a set of optimization tasks to generate multiple trajectories of self-optimization. The trajectories are filtered according to their optimization performance, and the better performing trajectory is selected. This optimization trajectory is used as the supervision signal to train the original Evolution Transformer model. After several iterations, it is hoped to find optimization strategies that are better than the original model.

These investigations primarily leverage the algorithm comprehension, representation, and generation capabilities of LLMs to enhance the performance of EAs. Rather than directly serving as search operators in the optimization process, LLMs are employed to refine EAs at the algorithmic level. This avenue holds significant promise, and in Section \ref{code_generation}, we will provide a comprehensive overview of code generation techniques based on LLMs and EAs, as well as their challenges, including handling intricate problem descriptions, processing large volume of numerical inputs, and addressing efficiency concerns stemming from time-intensive interactions.

\subsection{LLM Empowering EA with Other Capabilities}

In addition to using LLM as an evolutionary operator or directly generating optimization algorithms, LLM can also provide assistance for EAs from other aspects.
Chen \emph{et al.} proposed OptiChat \cite{chen2023diagnosing} to diagnose the infeasible model, identifying the main constraints that cause the infeasibility. Specifically, the infeasible model means that the constraints of an optimization algorithm cannot be satisfied simultaneously \cite{chinneck1991locating}, which are very common in practical applications, mainly because the model parameters are set incorrectly, or some conflicting new constraints are added. OptiChat uses LLMs to provide a natural language interface for non-expert users, making it more convenient to understand and repair infeasible optimization model problems. The user provides the definition of the optimization model, including decision variables, parameters, constraints, and objective function, by using the Pyomo algebraic modeling language \cite{hart2011pyomo}. OptiChat can provide explanations of the optimization model, identify potential sources of infeasibility, and offer suggestions to make the model feasible through interactive conversations.

Besides, Maddigan \emph{et al.} leveraged LLMs to provide explainability for results of EA, improving understanding for non-expert users \cite{maddigan2024explaining}. They focused on using GP for nonlinear dimensionality reduction, where GP individuals are tree expressions that directly map each new low-dimensional feature to combinations of high-dimensional features. Specifically, a system is constructed for running GP-based nonlinear dimensionality reduction processes and visualizing results. LLMs are integrated into the system to develop a conversational interface to provide explanations, using prompts containing dataset and tree expression information to provide contextual knowledge. They also built a vector repository utilizing vector similarity search to provide related literature knowledge to the model and address knowledge gaps. 
Another recent study uses LLM to help decision makers understand and interpret the optimal solution set in multi-objective evolutionary optimization \cite{singh2024enhancing}.



\section{EA-enhanced LLM}

This section delves into the emerging research of enhancing LLMs through the application of EA. The primary focus of this exploration is twofold: EA-based prompt engineering and EA-based LLM architecture search, as summarized in Tables \ref{tab:EA4LLM_table1} and \ref{tab:EA4LLM_table2}.

\subsection{EA-based Prompt Engineering}

Black-box prompt engineering \cite{diao2023blackbox} enables adjusting prompts without requiring access to the underlying model's parameters and gradients, making it particularly valuable for closed-source LLMs \cite{liang2023holistic}. This advancement allows for effective optimization of closed-source models, overcoming previous limitations imposed by the unavailability of model parameters. Currently, evolutionary computation plays a role in two types of prompt engineering: discrete prompt optimization and continuous prompt optimization. The former considers meaningful textual instructions as prompts, as shown in Section \ref{sec_prompt1}, while the latter uses vectors composed of continuous numerical values as prompts, as shown in Section \ref{sec_prompt2}.

\begin{table}[t]
\centering
\caption{Summary of EA-based Prompt Engineering.}
\resizebox{0.49\textwidth}{!}{
\begin{tabular}{|c|c|c|c|}
\hline
\textbf{Type} & \textbf{Method} & \textbf{Used EA} & \textbf{Ref.} \\
\hline
& GPS & GA & \cite{xu2022gps} \\
& GrIPS & Greedy search algorithm & \cite{prasad2023grips} \\
& EvoPrompt & GA, Differential EA & \cite{guo2023connecting} \\
& Plum & Metaheuristic algorithm & \cite{pan2023plum}  \\
Discrete & PromptBreeder & GA & \cite{fernando2023promptbreeder} \\
Prompt & SPELL & Any EA & \cite{li2023spell} \\
Optimization & EoT prompting & Any EA  & \cite{jin2024zero} \\
& iPrompt & Iteration similar to EA & \cite{singh2022iprompt} \\
& PhaseEvo & Designed framework  & \cite{cui2024phaseevo} \\
& InstOptima & NSGA-II  & \cite{yang2023instoptima} \\
& EMO-Prompts & NSGA-II, SMS-EMOA  & \cite{baumann2024evolutionary} \\ 
\hline
& BBT & CMA-ES & \cite{sun2022black} \\
& BBTv2 & CMA-ES & \cite{sun2022bbtv2} \\
Gradient-Free Soft & Clip-Tuning & CMA-ES & \cite{chai2022clip} \\
Prompt Optimization & Shen \emph{et al.} & CMA-ES & \cite{shen2023reliable} \\
& BPT-VLM & CMA-ES & \cite{yu2023black} \\
& Fei \emph{et al.} & CMA-ES & \cite{fei2023gradient} \\
\hline
Prompt Generation for & Evol-Instruct & Mutation and selection & \cite{xu2023wizardlm} \\
Data Augmentation & Sun \emph{et al.} & Any EA & \cite{sun2024dial} \\
\hline
& AutoDAN & GA & \cite{liu2023autodan} \\
Prompt Generation & Jailbreak Attacks & GA & \cite{lapid2023open} \\
for Security & SMEA & Any EA & \cite{zou2024system} \\
& Shi \emph{et al.} & Any EA & \cite{shi2024red} \\
\hline
\end{tabular}}
\label{tab:EA4LLM_table1}
\end{table}

\subsubsection{Textual Prompt Optimization}
\label{sec_prompt1}

Before the emergence of ChatGPT, several research studies had already explored the use of metaheuristic algorithms, such as EAs, for prompt engineering in pretrained language models. Notable examples include Gradient-free Instructional Prompt Search (GrIPS) \cite{prasad2023grips}, which employed a greedy search strategy, and Genetic Prompt Search (GPS) \cite{xu2022gps}, which utilized a Genetic Algorithm (GA) \cite{holland1992genetic}. These studies typically employed EAs as the underlying search framework, with the LLM being responsible for generating and evaluating prompts. However, it is important to note that these investigations were limited in their scope and primarily focused on specific prompt engineering scenarios.

In order to fully unlock the potential of discrete optimization in the realm of black-box prompt optimization, Guo \emph{et al.} proposed an EA-based discrete Prompt tuning framework (EvoPrompt) \cite{guo2023connecting}. In EvoPrompt, the LLM emulates evolutionary operators using a fixed prompt and generates new candidate prompts through crossover and mutation based on the differences between parent prompts. The EA is used to guide the optimization process and retain the best prompts. EvoPrompt has been validated using two evolutionary methods: GA and Differential EA \cite{storn1996minimizing}. Pan \emph{et al.} proposed a similar prompt optimization paradigm called Prompt Learning Using Metaheuristic (Plum) \cite{pan2023plum}, which formulates prompt learning as a black-box non-convex discrete optimization problem, allowing various metaheuristic algorithms to be applied to help discover effective prompts. Beside the GA \cite{holland1992genetic}, this work implemented another five algorithms, including Hill Climbing, Simulated Annealing \cite{kirkpatrick1983optimization}, Tabu Search \cite{glover1986future}, and Harmony Search \cite{geem2001new}. Their experiments showed that Harmony Search was more efficient than GA, achieving better performance with fewer API calls.

In contrast to EvoPrompt's initialization approach (hand-designed task-specific prompts), Fernando \emph{et al.} introduced PromptBreeder, an automated approach that utilizes LLM to optimize prompts based on problem descriptions \cite{fernando2023promptbreeder}. PromptBreeder employs an EA framework to automatically explore prompts in the problem domain and simultaneously evolves task prompts and mutation prompts, rather than using fixed prompts in EvoPrompt. Additionally, PromptBreeder adopts diverse mutation operators, including zeroth-order, first-order, and higher-order mutations, to more thoroughly explore the space of prompt strategies.
Cui \emph{et al.} proposed a prompt optimization framework called PhaseEvo that enables joint optimization of prompt instructions and examples \cite{cui2024phaseevo}. The framework adopts a four-phase optimization strategy and designs five different evolutionary operators to achieve different goals in each phase, including global initialization phase (using Lamarckian operator or semantic mutation operator to generate initial population), local feedback mutation phase (using feedback mutation operator to accelerate convergence), global evolution mutation phase (using EDA operator and crossover operator to escape local optima), and local semantic mutation phase (using semantic mutation operator to find the global optimum). Additionally, PhaseEvo uses performance-based vectors and Hamming distance to evaluate candidate similarity instead of commonly used semantic similarity. Other similar approaches include Semantic Prompt Evolution based on a LLM (SPELL) \cite{li2023spell}, zero-shot EoT prompting \cite{jin2024zero}, and interpretable autoprompting (iPrompt) \cite{singh2022iprompt}.

Moreover, Yang \emph{et al.} argued that 
the quality of instructions should not be measured solely from a performance perspective; other objectives such as instruction length and perplexity can be considered as well \cite{yang2023instoptima}. Similar to the aforementioned studies, Yang \emph{et al.} proposed InstOptima, which utilizes the ChatGPT to simulate instruction mutation and crossover operations. Under multi-objective optimization, InstOptima employs the NSGA-II algorithm \cite{deb2002fast} for non-dominated sorting and obtains a set of excellent instructions (the Pareto front) in terms of multiple objectives. EMO-Prompts proposed by Baumann \emph{et al.} also optimizes prompts from an evolutionary multi-objective optimization perspective \cite{baumann2024evolutionary}. It showcases an interesting scenario of finding prompts that cause the model to generate text containing two emotions.

\subsubsection{Gradient-Free Soft Prompt Optimization}
\label{sec_prompt2}

In some open-source LLMs, users can only access the model through the inference API and do not have access to the model parameters and gradient information. To optimize continuous prompts using limited samples in this scenario without relying on gradient-based optimization of model performance, Sun \emph{et al.} proposed Black-Box Tuning (BBT) \cite{sun2022black}. BBT represents the continuous prompts to be optimized as a vector with fixed initialization. It projects the prompt vector into a low-dimensional subspace using a random projection matrix to reduce the optimization difficulty. The CMA-ES is used to sample new values for this vector in the subspace, and the loss value of the vector is queried from the LLM to modify the distribution for the next iteration of CMA-ES. This iterative process continues until the stop condition is met, where the obtained subspace vector is then added to the fixed initialization vector and projected to obtain the final prompt vector.

Several studies aimed to improving the BBT. Sun \emph{et al.} proposed BBTv2 \cite{sun2022bbtv2}. BBTv2 adds continuous prompt vectors not only at the input layer but also at each hidden layer in LLM, forming a deep prompt representation. This significantly increases the number of tunable parameters, which is advantageous for handling more challenging tasks. To optimize higher-dimensional deep prompts, BBTv2 employs a divide-and-conquer algorithm to alternately optimize prompt vectors at different levels, decomposing the original problem into multiple lower-dimensional subproblems. Additionally, the random projection matrix is no longer generated using a uniform distribution but is adapted to different models based on the statistical distribution of model word vectors or hidden states, better accommodating different models. To improve search efficiency and provide more fine-grained and multi-perspective evaluation feedback, Chai \emph{et al.} replaced the method of reducing the space dimensionality by using a single random projection matrix in BBT with the Clip-Tuning approach \cite{chai2022clip}. Clip-Tuning applies Dropout sampling to the pre-trained model during inference, generating multiple sub-networks that can be viewed as projections of predictions for the samples in the original high-dimensional space. By blending rewards from multiple sub-network predictions, the search algorithm converges faster to the optimal solution. Shen et al. learned the complete distribution over soft prompts, rather than just considering a point estimate like BBT \cite{shen2023reliable}, in order to achieve uncertainty quantification of the predictive results. They used variational inference and ensemble learning to learn and approximate the posterior distribution over soft prompts. Similar to BBT, the EA in these studies is the CMA-ES. Some other studies focused on the vision-language models \cite{yu2023black,fei2023gradient}, which are not detailed here.

In addition to prompt engineering aimed at optimizing prompts, the prompt generation can have broader applications, such as data augmentation for training LLM or conducting jailbreak attacks on LLM to test security. Sections \ref{sec_prompt3} and \ref{sec_prompt4} will introduce how EA plays a role in these two tasks.

\subsubsection{Prompt Generation for Data Augmentation}
\label{sec_prompt3}

In contrast to the aforementioned approaches that purely focus on prompt optimization, Xu \emph{et al.} employed prompt data augmentation using LLM and EA, referred to as Evol-Instruct, to enhance instruction data \cite{xu2023wizardlm}. The framework uses a small number of manually written seed instructions as the initial dataset, treats LLM as an instruction evolver, iteratively evolves instructions based on different evolutionary prompts (e.g., adding constraints, deepening), and uses LLM to generate instruction responses and filter out ineffective instructions. The evolved instructions are then added to the training data, and the process is repeated iteratively to generate diverse and high-quality instruction sets. Experimental results demonstrate that Evol-Instruct improves the performance of instruction-based models on various downstream tasks, such as language modeling and text classification. Sun \emph{et al.} adopted a similar approach to build high-quality domain-specific instruction data, and designed a multidimensional quality evaluation method to assess the generated data \cite{sun2024dial}.

\subsubsection{Prompt Generation for LLM Security}
\label{sec_prompt4}


Another improvement of LLM focuses on model security. Specifically, alignment techniques ensure the safety and interpretability of model outputs by collecting human-annotated data and reinforcement learning methods \cite{ganguli2022red}, making the models generate responses that are more in line with human values and expectations. However, with properly designed problem prompts, known as ``jailbreak attacks", aligned LLMs can still generate inappropriate responses. Currently, handcrafted prompts for jailbreak attacks are difficult to automate and scale, while automatically generated prompts are often semantically incoherent and difficult to defend against semantics-based defenses \cite{jain2023baseline}. Therefore, some studies use EA-based prompt engineering for automatic jailbreak attacks. Liu \emph{et al.} model the jailbreak attack problem as an optimization problem and use EAs to automatically optimize prompts \cite{liu2023autodan}, namely Automatically generating DAN-series-like jailbreak prompts (AutoDAN). It utilizes existing handcrafted jailbreak prompts to initialize the population and designs an adaptive function suitable for structured discrete data like text. In addition, AutoDAN adopts a hierarchical GA to consider the optimization of prompts at the sentence and vocabulary levels, and designs a momentum word scoring mechanism to balance search ability and semantic coherence. 
The research by Lapid \emph{et al.} achieves a similar goal to AutoDAN \cite{lapid2023open}. Zou \emph{et al.} noted the importance of system messages in LLMs jailbreaking, as well as the transferability of jailbreaking prompts under different system messages. They proposed the System Message Evolutionary Algorithm (SMEA) \cite{zou2024system} to search for optimized system messages with stronger resistance against jailbreaking attacks. Another study focused on testing the robustness of reliability detection systems for LLM-generated texts \cite{shi2024red}. They utilized the protected auxiliary model ChatGPT to generate word substitution candidates, and performed query-based word substitution attacks based on EA. The search involved instructional prompts to change the generation style and make the detection system difficult to detect, which is essentially a prompt optimization process for adversarial purposes.

Based on the discussions above, EA-based prompt engineering has played a significant role in LLM. Researchers have used EAs as a search framework, combined with prompt generation and evaluation in both discrete and continuous prompt optimization. 
Additionally, EAs have also been applied to data augmentation and LLM security, further expanding the scope of prompt generation applications. However, current methods still face challenges such as the selection of initial populations, expanding search spaces, and method stability. Addressing these challenges will require better strategies for selecting initial populations, efficient search algorithms, and a deeper understanding of LLMs. Therefore, EA-based prompt optimization provides powerful tools and methods for improving LLM performance and expanding its applications. With further research and advancements, we can expect breakthroughs and innovations that will contribute to the progress of prompt optimization in LLM.

\subsection{EA-based LLM Architecture Search}
\label{sec_nasllm}

\begin{table*}[t]
    \centering
    \caption{Summary of EA-based LLM Architecture Search Methods}
    \label{tab:EA4LLM_table2}
    \begin{tabular}{|c|c|c|c|c|}
    \hline
    \textbf{Method}          & \textbf{Main Task}                             & \textbf{Used EA}    & \textbf{Evaluated LLM} & \textbf{Ref.} \\
    \hline
    AutoBERT-Zero          & Discover new universal LLM backbone from scratch & Any EA   & BERT        & \cite{gao2022autobert} \\
    SuperShaper            & Search hidden unit dimensions in each Transformer layer & Any EA   & BERT        & \cite{ganesan2021supershaper} \\
    AutoTinyBERT           & Automatically optimize LLM architecture hyperparameters & Any EA   & BERT        & \cite{yin2021autotinybert} \\
    LiteTransformerSearch & Independently vary hyperparameters of each decoder layer & Any EA   & GPT-2        & \cite{javaheripi2022litetransformersearch} \\
    Klein \emph{et al.}          & Various LLM architecture optimization methods     & Multi-objective EA                         & BERT                   & \cite{klein2023structural} \\
    Choong \emph{et al.}                   & Obtain specialized models optimized for specific tasks & MO-MFEA & M2M100-418M, ResNet-18              & \cite{choong2022jack} \\
    \hline
    \end{tabular}
\end{table*}

Prompt engineering goes beyond the realm of LLM models by optimizing the input format to enhance the quality of model outputs. Differing from prompt engineering, another approach known as LLM architecture search focuses on directly optimizing the architecture of LLM models to achieve superior performance and lighter LLM models. With the increasing complexity and size of neural networks, manual design and optimization of architectures have become laborious and time-consuming tasks. EAs offer an effective approach to automate the search process and discover promising architectures \cite{liu2021survey,zhou2021survey}. By employing evolutionary operators such as mutation, crossover, and selection, these algorithms can generate a diverse set of candidate architectures. This exploration-exploitation trade-off allows for a comprehensive exploration of the architecture space while gradually converging towards promising solutions. Previously, EA-based NAS methods were primarily applied to small-scale models and achieved promising results. Since the study by So \emph{et al.} focusing on NAS for Transformers \cite{so2019evolved}, EA-based NAS has been utilized on diverse large-scale models\footnote{The papers reviewed in this subsection mainly concentrate on designing NAS methods for pretrained language models (PLMs) without explicitly distinguishing between the concepts of LLMs and PLMs.} as shown in Table \ref{tab:EA4LLM_table2}.

The first work, using an NAS algorithm based on evolutionary search is AutoBERT-Zero \cite{gao2022autobert}, to discover a new universal LLM backbone from scratch. In AutoBERT-Zero, a well-designed search space is introduced, containing primitive math operations to explore novel attention structures in the intra-layer level, and leveraging convolution blocks as supplementary to attention in the inter-layer level. Additionally, this work proposed the Operation-Priority evolution strategy which utilizes prior information of operations to flexibly balance exploration and exploitation during search. Furthermore, a training strategy called Bi-branch Weight-Sharing is designed to speed up model evaluation by initializing candidates with weights extracted from a super-net. Different from the more complex search space in AutoBERT-Zero \cite{gao2022autobert}, SuperShaper proposed by Ganesan \emph{et al.} \cite{ganesan2021supershaper} focuses on searching the hidden unit dimensions of each Transformer layer. This is achieved by adding bottleneck matrices between each Transformer layer, thus enabling the variability of hidden dimensions across layers. By slicing the bottleneck matrices, different structured sub-networks are sampled from the super network, and the hidden dimensions between layers are determined by the slicing. The sub-network search process considers optimizing two objectives simultaneously, including perplexity and latency, and trained predictors are used to approximate them. Finally, an EA is used to search for the optimal hidden unit shape distribution across layers that simultaneously meets the requirements of accuracy and latency.

Some methods try to achieve NAS on more complex search space, where key hyperparameters of each layer of Transformer are considered, such as hidden state dimension, number of attention heads, and feedforward network dimension. Yin \emph{et al.} applied EA-based NAS methods for the first time to automatically optimize the architecture hyperparameters of LLMs \cite{yin2021autotinybert}. The proposed AutoTinyBERT searches within the space of structures with identical layer depths and dimensions, thereby simplifying the search space from exponential to linear scale and greatly reducing the search complexity. In addition, this work trained a big model SuperPLM with one-shot learning that contains all potential sub-structures. Therefore, when evaluating a specific structure, they do not need to train it from scratch, but can directly extract the corresponding sub-model from SuperPLM to serve as high-quality initializations for various latent structures. Yin \emph{et al.} used EAs to search, and designed a sub-matrix extraction method to quickly extract different structural sub-models from SuperPLM for evaluating structure performance, selecting elite structures and mutating to generate new generations, repeating this process to search for the optimal structure. LiteTransformerSearch proposed by Javaheripi \emph{et al.} \cite{javaheripi2022litetransformersearch} also allows the hyperparameters of each decoder layer, such as hidden size, number of heads and feedforward dimensions, to vary independently, creating a heterogeneous search space. The search space also includes other hyperparameters like number of layers and embedding dimensions. LiteTransformerSearch leverages the empirical observation that decoder parameter count has a high correlation with validation perplexity, establishing the first training-free low-cost proxy for Transformer architecture search - decoder parameter count. It uses EAs to sample candidate architectures from the search space based on this proxy, while also measuring hardware metrics directly on the target device. This enables LiteTransformerSearch to perform multi-objective NAS to obtain a Pareto frontier estimation that optimizes perplexity, latency and memory. And its effectiveness has been evaluated on two popular autoregressive Transformer backbones GPT-2 and Transformer-XL. Similarly, Klein \emph{et al.} also conduct NAS in multi-objective manner \cite{klein2023structural}, which discover multiple subnetworks of LLMs that balance model performance and size.

Different from the above methods, Choong \emph{et al.} did not directly study the NAS task, but proposed the concept of ``Set of Sets" \cite{choong2022jack}, which refers to a set of models that can simultaneously meet multiple task settings and resource constraints. They studied how to obtain a set of smaller-scale models optimized for specific tasks from LLMs through multi-objective multi-task EAs (such as the MO-MFEA algorithm \cite{gupta2016multiobjective,bali2020cognizant}). Among them, the LLM plays the role of a general-purpose basic model. Experimental results show that the specialized models obtained in this way can achieve better performance or greater compression rates than the original large model in various application fields and neural network architectures.

In conclusion, EAs have shown great potential in assisting LLM architecture search. By leveraging EAs, researchers can automate the process of optimizing LLM architectures, which is otherwise laborious and time-consuming. EAs enable the generation of diverse candidate architectures through evolutionary operators such as mutation, crossover, and selection, striking a balance between exploration and exploitation. Several studies have successfully applied EA-based NAS methods to different aspects of LLM architecture, including discovering new universal LLM backbones, optimizing hidden unit dimensions, and tuning hyperparameters. However, there are still some challenges remaining, such as the high time consumption and the limited generalization ability. Future research should focus on addressing these challenges and exploring more effective LLM architecture search approaches.

\subsection{EA Empowering LLM for Other Enhanced Capabilities}

In addition to enhancing the performance of LLM through NAS, some researchers have utilized the search capability of EA to assist in improving other aspects of LLM.

Before the emergence of LLM, Kim \emph{et al.} have studied the Length-Adaptive Transformer model \cite{kim-cho-2021-length}, which can automatically adjust the sequence length according to different computational resource constraints during inference. They proposed LengthDrop technique, which allows the model to randomly reduce the sequence length of each layer during training, making the trained model more robust to changes in sequence length during inference. On the trained model, an EA is used to continuously optimize the population of length configurations to maximize accuracy under various computational budgets. During inference, the corresponding optimal length configuration is directly used without retraining the model to achieve efficient inference. Jiang \emph{et al.} studied how to deploy generative inference services for large foundation models in a heterogeneous distributed environment to reduce the costs of centralized data centers \cite{jiang2023hexgen}. Their research allows each pipeline parallel stage to be assigned with different numbers of Transformer layers, while also allowing each stage to set different degrees of tensor model parallelism. They used a GA with operations of merging, splitting, and swapping different pipeline groups. By combining with a dynamic programming algorithm to evaluate the costs of each scheme, they searched for the optimal solutions of pipeline group allocation, GPU device allocation within each pipeline group, and layer allocation for each pipeline stage. Ding \emph{et al.} studied how to extend the context window of LLMs to 2048k tokens \cite{ding2024longrope}. They discovered two forms of non-uniformities in the positional encoding - across different dimensions and token positions. Through an EA, they searched for the optimal rescaling factors for each dimension and initial positions. This allowed different dimensions and positions to be rescaled with different degrees of interpolation or extrapolation. It achieved better preservation of the original positional information compared to existing methods that uniformly handled all dimensions. Akiba \emph{et al.} proposed a model merging approach that utilizes CMA-ES algorithm to optimize merged performance in both parameter space and data flow space \cite{akiba2024evolutionary}, which can automatically discover effective combinations of diverse models from different domains. Li \emph{et al.} propose to leverage a small model to store domain knowledge in order to assist LLMs in tackling the lack of domain-specific knowledge \cite{li2024blade}. To learn and apply the knowledge, the small model undergoes domain-specific pretraining, knowledge instruction tuning, and Bayesian optimization, where CMA-ES is used to find soft prompts that optimizes the consistency between the outputs of the two models. This allows the small model to better satisfy the needs of the large model in downstream tasks. Moreover, the studies of self-evolution in LLMs have also widely adopted the ideas of EAs, which has been discussed in \cite{tao2024survey}.

\section{Applications Driven by \\Integrated Synergy of LLM and EA}

In recent years, the synergy between LLM and EA has attracted increasing attention. Researchers combine the strengths of LLM and EA to enhance performance in various downstream applications, as shown in Table \ref{tab:application_summary}. This section provides a review from a problem-based perspective and discuss the collaborative effects of LLM and EA in several popular downstream applications.

\subsection{Code Generation}
\label{code_generation}

LLMs and EAs have both shown promise in automating code generation. LLMs can be trained on vast amounts of publicly available source code to gain a broad understanding of programming concepts and patterns \cite{chen2021evaluating,li2022competition}. However, their generation abilities are limited by their training data distribution. EAs, on the other hand, are capable of open-ended search through program spaces. But traditional mutation operators used in GP struggle to propose high-quality changes in a way that mimics how human programmers intentionally modify code. The joining of EAs and LLMs has opened up more opportunities for code generation.

\begin{table*}[t]
    \centering
    \caption{Summary of Applications Driven by Integrated Synergy of LLM and EA.}
    \label{tab:application_summary}
    \resizebox{\textwidth}{!}{
    \begin{tabular}{|c|c|c|c|c|}
        \hline
        \textbf{Category} & \textbf{Subcategory} & \textbf{Method} & \textbf{Description} & \textbf{Ref.} \\
        \hline
        \multirow{15}{*}{Code Generation} & \multirow{3}{*}{Universal Code Generation} & ELM & Universal method for code generation & \cite{lehman2023evolution,bradleyopenelm} \\
        &  & WizardCoder & Use Evol-Instruct to enhance the code generation & \cite{luo2023wizardcoder} \\
        &  & Pinna \emph{et al.} & Improve LLM-generated code by Grammatical Evolution & \cite{pinna2024enhancing} \\
        \cline{2-5}
        & \multirow{10}{*}{Domain-specific Code Generation}  & SEED & Data cleaning tasks & \cite{chen2023seed} \\
        & & EUREKA & Design reward in reinforcement learning & \cite{ma2023eureka} \\
        & & EROM & Design reward in reinforcement learning & \cite{narinevolutionary} \\
        & & Zhang \emph{et al.} & Design reward in reinforcement learning & \cite{zhangsimple} \\
        & & Diffusion-ES & Design reward in reinforcement learning & \cite{yang2024diffusion} \\
        & & GPT4AIGChip & Design AI accelerator & \cite{fu2023gpt4aigchip} \\
        && FunSearch & For mathematical and algorithmic discovery & \cite{romera2023mathematical} \\
        && L-AutoDA & For decision-based adversarial attacks & \cite{guo2024autoda} \\
        && LLM-SR & For scientific equation discovery from data & \cite{shojaee2024llm} \\
        && Shojaee \emph{et al.} & Node importance scoring functions in complex networks & \cite{mao2024identify} \\
        \cline{2-5}
        & \multirow{2}{*}{Security in Code Generation} & DeceptPrompt & Generates code containing specified vulnerabilities & \cite{wu2023deceptprompt} \\
        && G3P with LLM & Enhance code security & \cite{tao2023program} \\
        \hline
        \multirow{5}{*}{Software Engineering} & \multirow{2}{*}{Software Optimization} & Kang \emph{et al.} & Improves traditional genetic improvement & \cite{kang2023towards} \\
        &  & Brownlee \emph{et al.} & Improves traditional genetic improvement & \cite{brownlee2023enhancing} \\
        \cline{2-5}
        & \multirow{2}{*}{Software Testing} & TitanFuzz & Test case generation & \cite{deng2023large} \\
        &  &  CodaMOSA & Test case generation & \cite{lemieux2023codamosa} \\
        \cline{2-5}
        & \multirow{1}{*}{Software Project Planning} & SBSE & Optimize selection of example sets & \cite{tawosi2023search} \\
        \hline

        \multirow{8}{*}{Neural Architecture Search} & \multirow{1}{*}{Representation Capability of LLM} & GPT-NAS & Fine-tune GPT model to guide the NAS. & \cite{yu2023gpt} \\
        \cline{2-5}
        & \multirow{5}{*}{Generation Capability of LLM} & LLMatic & Generate code of architecture. & \cite{nasir2023llmatic} \\
        && Evoprompting & Architecture generation through soft prompt tuning. & \cite{chen2023evoprompting} \\
        && Guided Evolution & Use LLM to mutate and crossover the architecture code & \cite{morris2024llm} \\
        && Others & Use LLM as an operator to generate new architectures & \cite{zheng2023can, wang2023graph} \\
        && GPTN-SS & Employ GPT-4 to generate code of NAS algorithm &  \cite{zeng2024discovering} \\
        \cline{2-5}
        & \multirow{2}{*}{Reasoning Capability of LLM} & Jawahar \emph{et al.} & Performance predictors of architecture. & \cite{jawahar2023llm} \\
        && ReStruct & Use LLM as predictor and selector of structure & \cite{chen2024large} \\
        \hline
    \end{tabular}}
\end{table*}

\subsubsection{Universal Code Generation}

Lehman \emph{et al.} applied LLMs into GP, called evolution through large models (ELM) \cite{lehman2023evolution}. ELM utilizes a small number of initially hand-written programs as seeds, combining the MAP-Elites search algorithm \cite{mouret2015illuminating} with an LLM-based intelligent mutation operation to generate a large amount of functionally rich programs as design examples. This method improves the search efficiency of GP algorithms since the LLM trained on code promote a more intelligent and effective mutation operator and reducing the probability of random mutations generating meaningless code. Considering the LLM may achieve different performance improvements in different domains, these programs are used as a training set to train LLMs in order to guide the LLM to output better code. Furthermore, by splicing new conditional inputs onto the LLM, it can teach the LLM to generate conditionally. This model will be fine-tuned through reinforcement learning to make the model able to generate appropriate outputs based on observed conditions. Bradley \emph{et al.} proposed an open-source Python library based on ELM \cite{bradleyopenelm}. Luo \emph{et al.} applied the Evol-Instruct introduced in Section \ref{sec_prompt3} to generate a more complex and diverse training dataset \cite{luo2023wizardcoder} from the existing Code Alpaca instructions. They then fine-tuned the open-source code LLM StarCoder using this dataset, resulting in ``WizardCoder" with the state-of-the-art performance on code generation. Different from the above studies, Pinna \emph{et al.} did not directly apply evolutionary operators to code \cite{pinna2024enhancing}. The code generated by LLMs are used as the initial individuals. And representing these individuals as abstract syntax trees, Grammatical Evolution algorithm continuously improve the code individuals and ultimately obtains the optimal code.


\subsubsection{Domain-specific Code Generation}

The combined approach of using LLM and EA has demonstrated practicality in various domains. These methods essentially leverage the iterative search framework of EA and the code generation and text understanding capabilities of LLM, with LLM acting as an evolutionary operator in the code evolution process. By iteratively improving the code, these methods enable the generation of algorithmic code by utilizing prompts that typically include contextual information such as environment source code, task descriptions, or example code. This approach has been applied to generate code for a wide range of tasks, including customized data cleaning methods \cite{chen2023seed}, reward function design in reinforcement learning \cite{ma2023eureka,narinevolutionary,zhangsimple,yang2024diffusion}, automated AI accelerator design \cite{fu2023gpt4aigchip}, mathematical and scientific discoveries \cite{romera2023mathematical,shojaee2024llm}, decision-based adversarial attacks \cite{guo2024autoda}, and criticality evaluation of nodes in complex networks \cite{mao2024identify}.

In these research studies, additional innovations have further enhanced the collaborative effect of LLM and EA. For example, as introduced by Ma et al., the Evolution-driven Universal REward Kit for Agent (EUREKA) tracks the changing values of each component in the reward function throughout the reinforcement learning training process \cite{ma2023eureka}. It monitors metrics like the value curve of an individual penalty term over time. EUREKA generates a list of these tracked changes in a reward feedback text provided to the language model for reference. This allows the method to guide the language model towards targeted improvements to the reward function design. Additionally, EUREKA also permits humans to provide purely textual descriptions of the strengths and weaknesses of the current reward function's behavior as well as directions for desired behavioral changes. It takes these suggestions as context for the next round of search, generating reward functions that are better aligned with human preferences \cite{ma2023eureka}. In Narin's work, they additionally utilize the capabilities of a multimodal LLM, GPT-4V, to comprehend the visual information in the environment and assist in the generation process \cite{narinevolutionary}. The FunSearch method designed by Romera \emph{et al.} uses an island-based EA strategy to split the program database into multiple subpopulations that evolve in parallel \cite{romera2023mathematical}. This can effectively avoid local optima and maintain exploration diversity. Moreover, clustering programs based on their signatures can effectively identify programs with similar functionality but different code implementations, thus preserving more diverse programs. This balancing of exploitation and exploration enhances the algorithm's ability to more comprehensively search the solution space and discover globally optimal solutions. The GPT for AI Generated Chip (GPT4AIGChip) proposed by Fu \emph{et al.} introduces a demo-augmented prompt generator \cite{fu2023gpt4aigchip}. This generator can automatically select the two most relevant demonstrations from the demonstration library that correspond to the input design instruction. It augments the selected demonstrations into the prompt as contextual information to effectively guide the LLM towards generating more accurate code. The method proposed by Mao \emph{et al.} optimized various details including population design, management, evaluation, and experimental design \cite{mao2024identify}.

\subsubsection{Security in Code Generation}

In addition to directly generating code, some studies have focused on the security of generated code. Wu \emph{et al.} concerned about how to evaluate whether the generated code is vulnerable to attacks \cite{wu2023deceptprompt}. They proposed DeceptPrompt, which can generate adversarial natural language prefixes/suffixes to drive code generation models to produce functionally correct code containing specified vulnerabilities. DeceptPrompt continuously optimizes the prefixes and suffixes using GA, with LLM serving as the mutation operator. The fitness function is separately designed for the functionally correct part and the vulnerable part to guide the generated code to both retain functionality and contain the specified vulnerabilities. Empirical studies have shown that DeceptPrompt can successfully attack various popular code generation models. Tao \emph{et al.} further considered methods to enhance the security of generated code \cite{tao2023program}. They proposed a simple idea of combining LLM with Grammar-Guided GP (G3P) system \cite{tao2022multi}, which enforces that any synthesized program adheres to Backus-Naur form (BNF) syntax, thus promoting the development/repair of incorrect programs and reducing the chances of security threats. Specifically, LLM is primarily employed to generate the initial code population in G3P, where the initial population is mapped to a predefined BNF syntax program before evolutionary search, allowing for improvements of these programs using program synthesis benchmarks. Afterwards, these programs will be handed over to G3P and evolved into better code.

\subsection{Software Engineering}

Due to the promising performance of LLM and EA in code generation tasks, some studies have further applied them to practical problems in software engineering \cite{fan2023large,yan2023coco}, including subtakss software optimization, software testing, and software project planning.

\subsubsection{Software Optimization}

Genetic Improvement (GI) \cite{petke2017genetic} is a technique for automatically software optimization based on the ideas of evolutionary computation. Kang \emph{et al.} leverage the advantages of LLM in code understanding and generation to improve the shortcomings of blind mutation in traditional GI \cite{kang2023towards}. The usage is similar to the code generation steps in Section \ref{code_generation}, firstly letting the LLM improve the code for specific optimization objectives (such as time efficiency and memory consumption), then injecting the mutations generated by LLM into the candidate pool of GI, and continuously performing evolutionary operations until termination to obtain optimized code. Similarly, another study also enhanced the mutation operators of GI by leveraging LLMs to improve non-functional properties or fix functional bugs of software \cite{brownlee2023enhancing}. Brownlee \emph{et al.} believed that the search space of GI is limited by the mutation operators it uses. By treating LLMs' suggestions as additional mutation operators, they aimed to enrich the search space and thus obtain more successful mutation results (generated code). 

\subsubsection{Software Testing}

EAs have extensive applications in software testing techniques, where they are employed to search for effective test cases within the testing space to uncover errors and defects in software systems \cite{fraser2011evolutionary,fraser2011evosuite}. Recently, some studies combine the code generation capability of LLMs and search capacity of EAs, and apply them to software testing. Deng \emph{et al.} proposed TitanFuzz \cite{deng2023large} to test for bugs in deep learning libraries. The core idea of TitanFuzz is to directly leverage LLMs that have been pretrained on tens of billions of code snippets, which implicitly learned the syntax and API constraints. It utilizes the generative ability of Codex \cite{chen2021evaluating} for seed generation and the infilling ability of InCoder \cite{fried2022incoder} for mutation generation, to automatically produce a large number of input programs that meet the requirements for testing. The fitness function considers both the depth of the dataflow graph obtained from static analysis and the number of unique APIs called in the program, as well as the number of repeated API calls (as a penalty). This encourages the generation of programs with more complex API usage and richer API interactions. 
Lemieux \emph{et al.} took a more direct approach \cite{lemieux2023codamosa}, namely CodaMOSA. They first used conventional evolutionary search Many-Objective Sorting Algorithm (MOSA) \cite{panichella2017automated} and monitored its coverage progress. When the algorithm reached a coverage plateau (state where further mutation of test cases finds it difficult to improve coverage \cite{aleti2017analysing,albunian2020causes}), it would identify functions with low coverage in the module, and use these low-coverage functions as hints to request the LLM (e.g., CodeX \cite{chen2021evaluating}) to generate test cases. Afterwards, on the basis of the test cases generated by the LLM, the EA continued exploration. This method can significantly improve the coverage of test samples.

\subsubsection{Software Project Planning}

Story points are a commonly used method in software project planning and cost estimation to gauge the amount of work required to complete a user story \cite{tawosi2022agile}. One way to leverage LLMs for this task is through few-shot learning, by providing some labeled examples of past user stories and their estimated points to improve the model's ability to estimate new stories. However, the choice and number of examples can impact the estimation effectiveness. Tawosi \emph{et al.} propose Search-Based Optimisation for Story Point Estimation (SBSE) to optimize the example selection through a multi-objective approach \cite{tawosi2023search}. They use NSGA-II to simultaneously optimize three objective functions - the sum of absolute errors, the confidence interval of the error distribution, and the number of examples. This searches for the Pareto front of user story example sets. The GA explores the trade-off between estimation accuracy and complexity, as measured by the objective functions. The optimized Pareto front provides decision makers with different accuracy-complexity choice options when selecting an example set for few-shot learning with LLMs to estimate story points.

\subsection{Neural Architecture Search (NAS)}

LLMs and EAs also contribute to another field known as NAS\footnote{Methods reviewed in this section differ from those presented in Section \ref{sec_nasllm}. Approaches discussed in Section \ref{sec_nasllm} primarily focus on LLM architecture search, and their techniques are based on EAs, whereas methods reviewed in this section leverage the synergistic combination of EAs and LLMs. Moreover, these NAS methods are more versatile and not limited to LLM architecture search alone, applicable to a broader range of NAS tasks.}. Over the years, EAs have found widespread application in NAS \cite{liu2021survey,zhou2021survey}, revolutionizing the way we design and optimize neural networks. Their ability to simulate natural selection and iteratively improve architectures has proven invaluable in the quest for efficient and high-performing models. However, as we delve into the realm of cutting-edge advancements, a new player has emerged onto the NAS scene - LLMs. With their immense computational capacity and deep understanding of language and context, LLMs bring a fresh perspective to NAS. They possess the potential to offer novel insights and innovative solutions, empowering researchers and engineers to explore uncharted territories of network design. In the NAS applications based on LLM and EA, EAs are commonly employed to establish effective search frameworks, while LLMs leverage their unique abilities to contribute to NAS from diverse angles, including their representation capability, generation capability, and abundant prior knowledge.


\subsubsection{With Representation Capability of Fine-tuned LLM}

Yu \emph{et al.} proposed a NAS method called GPT-NAS \cite{yu2023gpt} that utilizes a GPT model \cite{radford2018improving} to guide the search of neural architectures. In GPT-NAS, neural architectures are encoded as inputs for the GPT model using a defined encoding strategy. Then the GPT model is pre-trained and fine-tuned on neural architecture datasets to introduce prior knowledge into the search process. In the architecture search, GPT-NAS uses an EA with operations like crossover and mutation to optimize individuals and obtain offspring with better performance. The fine-tuned GPT model is then used to effectively predict excellent new architectures based on previous structural information, thereby reconstructing the sampled architectures to guide and optimize the entire search process. In this method, the EA and GPT achieve complementary advantages. The EA excels in search and optimization, yet its direct application may face limitations due to constraints in variation capability and the vastness of the search space. This can pose challenges in effectively uncovering truly exceptional architectures. By reorganizing the structures on the basis of individuals evolved by GPT, richer prior knowledge can be introduced to optimize the network architectures, effectively reducing the actual search space and guiding the search towards high-performance areas.

\subsubsection{With Code Generation Capability of General LLM}

Nasir \emph{et al.} leveraged the code generation ability of large models to help NAS \cite{nasir2023llmatic}. The proposed LLMatic does not directly search in the structure representation space, but uses the code of neural network structures as search points. Specifically, LLMatic establishes two archives: a network archive that stores trained network models, and a prompt archive that stores prompts and temperature parameters for generating networks. LLMatic employs QD algorithms \cite{pugh2016quality} to perform the search, utilizing the repeated process of generation, training and selection to continuously optimize the quality and diversity of networks. The large model CodeGen \cite{nijkamp2022codegen} is used to complete crossover and mutation in the evolutionary process: in the mutation operation, CodeGen makes subtle modifications to the selected network based on the mutation prompt. In the crossover operation, CodeGen partially matches the selected multiple networks based on the crossover prompt. Chen \emph{et al.} took a similar approach by using the PALM \cite{chowdhery2023palm} as the crossover and mutation operators for the NAS task in the evolutionary search process. Moreover, they further improved the LLM's ability of generating candidate architectures through soft prompt tuning during NAS \cite{chen2023evoprompting}. The Guided Evolution framework proposed by Morris \emph{et al.} also uses LLM to mutate and crossover the code \cite{morris2024llm}. The main EAs adopted are SPEA-2 \cite{zitzler2001spea2} and NSGA-II. SPEA-2 is used to retain elite individuals that performed well in the previous generation. NSGA-II is used to select individuals for crossover and mutation. At the same time, they proposed ``Evolution of Thought'' to continuously optimize the recommendations of LLM, which allows the LLM to propose improved modifications for new individuals based on the performance of past mutations, making the evolutionary process more data-driven and efficient. In addition to the above study, some papers \cite{zheng2023can,wang2023graph} do not explicitly mention using an EA framework, but employ LLM as an operator to generate new architectures. In addition to using LLM for architecture code generation, GPTN-SS employed GPT-4 to assist in generating code for a tensor network structure search algorithm \cite{zeng2024discovering}. Similar to other code generation methods discussed in Section \ref{code_generation}, EA provides an iterative optimization framework, while LLM is utilized for evolutionary operations on the code.

\subsubsection{With Reasoning Capability of LLM Based on Its Abundant Prior Knowledge}

Unlike previous works that used LLM in the NAS search process, Jawahar \emph{et al.} utilized LLM to build performance predictors \cite{jawahar2023llm}. Specifically, they designed predictor prompts (including task description, architecture parameter definitions, examples, etc.) to define the prediction task. The prompts and test architectures would be input into the LLM (such as GPT-4), and the LLM could make predictions based on the architecture knowledge it learned during pre-training by understanding the complex relationships between architectures and performances. Since LLMs learned architecture-related knowledge from a large amount of literature, and this type of predictor is designed simply without needing to train an over-parameterized network, the deployment cost is low. Chen \emph{et al.} designed a novel reasoning meta-structure search framework called ReStruct to automatically discover meta-structures in heterogeneous information networks \cite{chen2024large}. They developed a grammar translator that encodes meta-structures into natural language sentences, which facilitates LLMs' understanding of structural patterns. Based on this, they designed a predictor and selector based on LLMs. The predictor adopts the idea of few-shot learning - it randomly samples some instances from the historical structure-performance records stored in previous searches, and uses these instances to prompt the LLM to predict the performance of new generated candidate structures. The selector will receive the prediction results of various candidate structures given by the predictor, and interact with the LLM to obtain selection suggestions. These modules are used in the selection process of GA, and combined with the designed insertion, grafting, and deletion operations to continuously optimize the meta-structure population and drive the search process. ReStruct also includes an explanation module based on chain-of-thought prompting, which can ultimately obtain semantically explainable and high-performance meta-structures.

\subsection{Other Generative Tasks}

In addition to the three widely applied scenarios mentioned above, the collaboration between LLM and EA has also driven the performance improvement of more generative tasks. This subsection provides a brief introduction to these studies. Most of these studies combined the generative ability of LLM and the search ability of EA, generating better results based on rich prior knowledge and text understanding.

\subsubsection{Text Generation} Text generation is one of the most direct applications of LLMs. Xiao \emph{et al.} used the generation ability of LLM and the search ability of EA for news summary generation \cite{xiao2023enhancing}, where event patterns structure is employed to represent and describe the key information, relationships and characteristics of events. They used LLM to extract event patterns from text, and then used GA to evolve the event pattern pool according to the importance and information quantity of argument roles, selecting the event pattern with the highest fitness to input into LLM for news summary generation. Some studies draw inspiration from text generation to optimize non-textual domains. Lim \emph{et al.} developed the SCAPE system to explore conceptual architecture design by combining EA and LLM \cite{lim2024scape}. SCAPE represents design schemes or concepts using text, which allows EA to endow LLM with stronger innovative capabilities. The system enables architects to participate in parent selection, significantly improving the quality and exploration efficiency compared to vanilla generative AI tools. Another interesting application is to generate Super Mario game levels using LLMs \cite{sudhakaran2023prompt,MarioGPT2023illuminating}. Sudhakaran \emph{et al.} used a string representation method similar to the Video Game Level Corpus dataset \cite{summerville2016vglc} to encode levels, and then trained a LLM namely MarioGPT based on GPT-2 that can generate Mario levels according to natural language descriptions. By combining MarioGPT with the Novelty Search algorithm \cite{lehman2008exploiting}, an open-ended level generation process was formed. Novelty search can continually discover levels with diverse structures and gameplay styles.

\subsubsection{Text-to-image Generation} In this type of task, when using LLMs like Stable Diffusion \cite{chefer2023attend} to generate photorealistic images, it is necessary to optimize the input text prompts and model parameters in order to achieve better performance. Berger \emph{et al.} tried to solve this problem and proposed the StableYolo method \cite{berger2023stableyolo}. Its main innovation lies in combining EAs with computer vision for the first time, simultaneously optimizing language prompts and model hyperparameters to improve image generation quality. Among them, the LLM Stable Diffusion plays the main role of image generation, generating a series of images based on the input text prompts. While the GA plays the role of searching and optimizing. It uses the recognition confidence given by the YOLO object detection model as the fitness function to perform multi-objective optimization. The optimization objects include keywords in text prompts and various hyperparameters of the Stable Diffusion model. Through repeated iteration, the best language prompts and model settings that maximize image quality are found.

In addition to text-related tasks, some applications in natural sciences and social sciences are provided as follows to demonstrate the combined advantages of LLMs and EAs.

\subsubsection{Natural Science} 
The researchers from McGill University conducted experiments using GPT-3.5 to fragment and recombine molecules \cite{jablonka202314} represented as a simplified molecular-input line-entry system (SMILES) string \cite{weininger1988smiles}. The LLM was able to successfully fragment molecules at rotatable bonds with a $70\%$ success rate. When recombining molecules, the LLM generated new combined molecules with fragments from each parent molecule that were chemically reasonable more often than a random recombination operation. The researchers also explored using the LLM to optimize molecular properties based on a similarity score to vitamin C. By providing the LLM with a set of molecules and their performance scores, the LLM generated new molecules that it believed would improve the score. The generated modifications were found to be more chemically sound compared to traditional GAs. Teukam \emph{et al.} utilized LLM and GA to optimize enzyme sequences, thereby improving their catalytic activity and stability \cite{teukam2024integrating}. Specifically, a protein language model based on evolutionary scale modeling (ESM-2) was employed to generate mutations and predict their functions. GAs then optimized the sequences through iterative selection and crossover operations, enhancing the predicted functionality. Ultimately, this approach yielded enzymes with optimized sequences that demonstrated enhanced catalytic performance compared to their wild-type counterparts.

\subsubsection{Social Science} Suzuki \emph{et al.} applied LLMs and EAs to study the evolutionary dynamics of social populations under conflicts of interest and behaviors between individuals \cite{suzuki2023evolutionary}, i.e. the propagation and changes of different personality traits and behavioral strategies. They used LLMs to make trait expressions more complex and higher-order, able to directly describe personality and psychological attributes that are difficult to model directly, and map them to behaviors. They then constructed an agent-based model where each agent uses a short natural language description to represent a personality trait related to cooperation as its gene. The LLM was used to derive a behavioral strategy for each agent based on the personality trait description and game history. The population evolved according to a GA, with offspring inheriting mutated versions of their parents' trait genes, thus simulating the evolutionary dynamics of population changes over generations. This study provided a novel approach for researching the evolution of complex social populations.


In summary, the collaboration between LLMs and EAs has driven advancements in distinct generative task domains. By exploiting their strengths, researchers have made significant progress in these domains, paving the way for further advancements in generative modeling and related fields.


\section{Roadmap and Future Directions}

\begin{figure}[t]
\begin{center}
\includegraphics[width=0.48\textwidth]{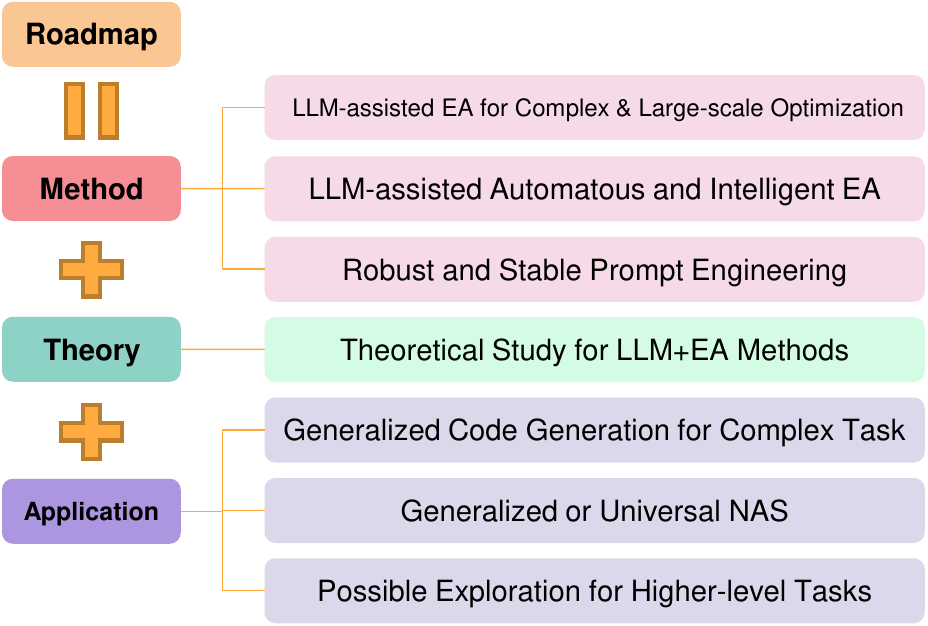}
\end{center}
\caption{Roadmap and Future Directions.}
\label{Roadmap}
\end{figure}

In the previous sections, we have reviewed the recent advances in unifying LLMs and EAs, but there are still many challenges and open problems that need to be addressed. In this section, we discuss the future directions of this research area. As depicted in Fig. \ref{Roadmap}, our perspective for future research includes investigations into methods, theory, and applications. Some of these studies aim to build upon the strengths and weaknesses highlighted in the reviewed research to propose more advanced approaches. Meanwhile, other research endeavors seek to explore new frontiers that go beyond the current body of knowledge, aiming to forge new research questions and infuse the field with renewed energy. This section will provide a separate introduction to these future research directions.

\subsection{LLM-assisted EA for Complex \& Large-scale Optimization}

Existing research has partially validated the ability of LLMs to address small-scale numerical optimization problems \cite{yang2023large,meyerson2023language,liu2023largeb}, but practicality still poses challenges. Firstly, for complex optimization problems with high-dimensional search spaces, numerous constraints, and high-precision numerical optimization, the limited context understanding and prompt length restrictions of large models may increase the difficulty of interaction with LLMs. Secondly, as black-box optimizers, the decision-making process of LLMs is difficult to interpret \cite{saha2023large}, and it remains unknown whether their optimization ability stems from reasoning or randomness. Moreover, the scope of the evaluated problems in existing research is relatively narrow \cite{yang2023large,meyerson2023language,liu2023largeb}, often limited to specific optimization problems, and the assessments only consider limited influencing factors, which are insufficient to fully demonstrate the optimization capabilities of LLMs. Another issue is that empirical studies suggest that LLMs may struggle to handle constrained problems effectively, making it crucial to incorporate constraint conditions into LLM optimization frameworks \cite{huang2024exploring}.

\todo{Therefore, to further explore and utilize the capabilities of LLMs in solving optimization problems, future work can focus on addressing the aforementioned challenges and improving various aspects such as LLM behavior interpretation and evaluation, rational utilization of optimization information, and solving complex problems. Currently, efforts can be made to interpret and understand the behavior of LLMs, such as analyzing their internal attention mechanisms \cite{jain2019attention} or investigating the relationship between LLM temperature settings and exploration of optimal solutions, in order to gain a more objective understanding of LLM optimization abilities. Based on these foundations, improvement strategies or new pre-training models can be proposed to solve optimization problems with higher dimensions and complex constraint conditions. Beyond these research directions, the contextual understanding and processing abilities of LLMs can be further exploited, for example, by analyzing historical information of the optimization process and considering more advanced prompt methods \cite{liu2023pre}. Conducting validation on large-scale, real-world problems going forward would also provide meaningful value.}

\subsection{LLM-assisted Automatous and Intelligent EA}

LLM holds significant promise for broader application scenarios within evolutionary optimization. In the following, we introduce several prospective areas, which, despite being relatively underexplored at present, are expected to drive more automatous and intelligent EAs in black-box optimization across diverse domains.

First, multi-modal LLMs provide opportunities for cross-domain EAs. Some LLMs have learned the relationships between different modalities during pre-training, allowing them to be used for downstream tasks involving multiple modalities \cite{wu2023multimodal,cui2024survey}, such as image captioning and text-to-image generation. The ability of LLMs to understand correspondences across modalities can help EAs achieve cross-modal crossover and mutation by providing cross-modal prompts or examples. Additionally, population construction and evaluation can be improved across modalities. Furthermore, as LLMs technologies continue to progress, evolutionary operators based on LLMs will also benefit and improve the performance of EAs. With increasing model scales and stronger generalization capabilities, evolutionary operators based on LLMs will be better able to simulate more complex evolutionary mechanisms precisely. This provides potential for tackling problems in complex search spaces. The emerging and rapidly advancing field of LLMs will also provide more choices for EAs. LLMs with different pre-training characteristics will continue to enrich and enhance the capabilities of evolutionary operators based on LLMs. 

In addition, LLM is trained on a large corpus of textual data and encompasses knowledge from various domains. It can serve as a knowledge base to assist evolutionary computation in better integrating domain knowledge, thereby improving optimization efficiency or optimality. For example, LLM's domain knowledge can be utilized to provide good initial solutions or improve problem formulation, including solution encoding, definition of solution space, and more. LLM can also provide valuable design principles for algorithm design based on its knowledge, enabling EC to tackle complex problems such as multi-objective, discrete, and dynamic problems through knowledge transfer.

\subsection{Robust and Stable Prompt Engineering}

One of the primary approaches to improving LLM through EAs is prompt engineering. Currently, a common method involves using LLM as an evolutionary operator to continuously generate new prompts within an evolutionary framework \cite{guo2023connecting,fernando2023promptbreeder,li2023spell}. This approach has been proven effective and superior in numerous research studies. However, several challenges still remain. Firstly, the initial population of the evolution process significantly influences the results. This is because a general and adaptable prompt template plays a crucial role in prompt generation and efficacy \cite{chen2023evoprompting}. Random initialization may struggle to leverage prior knowledge, while manual initialization can introduce bias. Additionally, for problems with abundant prior information, the prompt search space can be extensive. As the prompt length and vocabulary space increase, the search space exponentially grows, which can lead to overfitting or getting stuck in local optima. Lastly, these methods lack stability and heavily depend on the capabilities of LLM \cite{li2023spell}, making them susceptible to randomness. If LLM fails to comprehend and effectively utilize prompts, the effectiveness of the methods may be limited.

Regarding the issue of initialization, future research might consider multi-source initialization, simultaneously utilizing LLM to automatically expand the size and quality of the initial population. For problems with a high search space complexity, it is necessary to design more efficient evolutionary strategies, such as introducing richer evolutionary operators, leveraging the strengths of different EAs, and employing adaptive evolution. Another promising direction worth exploring is the utilization of human feedback to expedite and optimize the prompt discovery process, which has been overlooked in current research. In addressing the constraint of LLM on prompt performance, apart from considering the use of more powerful LLM models or designing dedicated fine-tuning methods, an alternative research direction at a lower level is to extract internal representations of LLM and study their mechanisms for understanding prompts, thereby guiding the design and representation of prompt strategies more effectively. Stable prompt optimization is also a promising direction to mitigate performance fluctuations caused by randomness.

\subsection{Theoretical Study for Specific LLM+EA Methods}

Although empirical studies have validated the effectiveness of combining LLM and EA on small-scale problems, the incentives of their interaction are not yet clear. This reminds us to explore the sources of mutual promotion between LLM and EA in theoretical research, and to analyze in detail their complementary advantages and existing issues in large-scale empirical studies, thereby further promoting further improvement. This research can delve into two aspects:

\subsubsection{Algorithm Analysis}
Researchers can analyze specific algorithms that combine LLM and EA and explore their convergence, complexity, and other properties. For convergence analysis, researchers can verify whether the algorithm can converge to the optimal solution or local optimum during the iterative process. This may involve proving the optimization properties of the objective function or fitness function during the algorithm's iterative process and analyzing the impact of algorithm parameter settings on convergence. Regarding complexity analysis, future work can analyze the time and space complexity of algorithms that combine LLM and EA. This includes evaluating the complexity of key steps in the algorithm, such as generating new individuals, selection, crossover, and mutation operations. The applicability of methods that combine LLM and EA on different types of problems is another perspective could be studied, which may involve analyzing the relationship between problem attributes, constraints, problem size, and algorithm performance, as well as studying the performance guarantees or theoretical limits of algorithms on specific problem types.

\subsubsection{Optimization Theory}
In problem modeling and analysis, specific problems that combine LLM and EA can be modeled and analyzed from the perspective of optimization theory. This may involve defining and characterizing the objective function, constraints, and exploring the feasible solution space of the problem. The key point of research lies in the search strategy, where researchers can study how to select and design appropriate search strategies to improve the performance of LLM and EA combination. This may include comparing and analyzing the effects of different search strategies (such as selection, crossover, and mutation operations) and sensitivity analysis of search strategy parameters. Moreover, future work can analyze the complexity of these methods on specific problems, including analyzing the computational complexity category of problems (such as P, NP, NP-hard) and analyzing the approximate performance of algorithms to determine the theoretical limits of algorithms in solving specific problems.



\subsection{Generalized Code Generation for Complex Task}
\label{LE_code_generation}

LLM and EA jointly contribute to another field of advancement, i.e., code generation. Currently, a plethora of research has emerged in this domain, which has further spurred the development of various downstream tasks, such as software engineering and EA design.

In code generation methods, one approach involves utilizing LLM to generate a substantial amount of training data and subsequently employing reinforcement learning to fine-tune LLM \cite{lehman2023evolution}. However, a significant challenge with such methods lies in the diversity and scale of the training data. LLM's training process heavily relies on extensive data support, and the LLM+EA generation approach may not cover all possible use cases. Moreover, the code generated by LLM might converge to a single solution, thereby limiting the model's capacity for generalization \cite{lehman2023evolution}. To address these limitations and generate more diverse training data that can adapt to a broader range of domains, one potential improvement is to incorporate continuous learning mutation operators. These operators can track changes in the problem space and introduce multiple initial solutions, facilitating multiple restarts of the search process. Additionally, by recording the mutations produced by each mutation operator and providing rewards or penalties based on the performance of the mutated offspring, the operators can also learn higher-quality mutation patterns.

Another approach type focuses on leveraging LLM's code generation capabilities and EA's search framework to continuously enhance code generation \cite{chen2023seed,ma2023eureka,fu2023gpt4aigchip,romera2023mathematical}. However, these methods encounter difficulties when dealing with complex algorithmic logic \cite{chen2023seed}. For tasks that involve intricate logic, a single code snippet may prove inadequate, necessitating the collaboration of multiple code snippets. Unfortunately, LLM's performance in automatically generating such code collections is suboptimal, and the existing limitations on LLM's input-output length make it challenging to handle large-scale and complex logic code \cite{chen2023seed}. To mitigate these challenges, one possible approach is to modularize complex algorithmic logic by designing a universal modular design and generation method that can decompose complex tasks. Alternatively, a more user-friendly and straightforward approach could involve developing an interactive interface that allows users to determine how tasks should be decomposed. Subsequently, LLM and EA can generate code for each subtask accordingly.

\subsection{Generalized or Universal NAS}

In the realm propelled by the collaboration of LLM and EA, NAS stands out as a vital application scenario. The prior knowledge about network architecture and code generation capability of LLM can greatly assist EA in efficiently discovering optimal architectures. However, existing work still faces various challenges. Some of these challenges are common among NAS methods based on EAs, such as high time consumption \cite{ren2021comprehensive}. Additionally, the integration of LLM introduces new issues. Firstly, it is important to note that the current state-of-the-art LLM models, while excelling in various tasks, are not specifically tailored for NAS. Their ability to tackle NAS tasks stems from the presence of network architecture-related information in the training data \cite{yu2023gpt}. Consequently, different LLM models exhibit significant variations in performance in NAS tasks. Furthermore, when compared to mainstream NAS methods, LLM-based approaches still exhibit some gaps in terms of their application scope and generalization ability. Additionally, certain ablation experiments reveal that the direct utilization of LLM prompts yields suboptimal learning results \cite{chen2023evoprompting}, underscoring the limitations of the current capabilities of LLM. Exploring fine-tuning methods that combine LLM with EAs has proven necessary.

Addressing these aforementioned issues would enhance the joint performance of LLM and EA in NAS tasks, unlocking the untapped potential of LLM in NAS and accelerating EA's search speed. Firstly, it is crucial to evaluate the performance of different LLM models in NAS tasks. Conducting fair evaluation experiments would validate the application scope and generalization ability of these LLM models across diverse network architectures. Leveraging additional training data to augment LLM's NAS capabilities and mitigating the impact of LLM pretraining quality on search results would improve robustness and stability. Moreover, optimizing the deeper structure of LLM during the fine-tuning process holds promise in generating superior solutions. To address efficiency concerns, subsequent research should explore leveraging historical search knowledge to expedite future searches and provide well-defined search spaces for LLM.

\subsection{Applications and Innovations in Higher-level Tasks}

\todo{In the discussed techniques within this paper, we observe their role as submodules in higher-level tasks. For instance, prompt optimization plays a role in LLM-based tasks, while code generation serves a purpose in NAS or software engineering tasks. In fact, these techniques can be utility in a broader range of higher-level tasks. Taking software engineering as an example, in higher-level tasks (such as repository-level software engineering), LLM and EA can fulfill more applicable demands. LLM can be used to implement cross-project collaboration. LLM not only can analyze the source codes and histories of multiple related projects to provide a large-scale search space for EA, but also its inherent prior knowledge can be used to identify the correspondence between knowledge of different projects, where EA can then utilize this relationship to perform knowledge transfer. In addition, EA can automatically generate version control strategies for different development stages, such as how to merge branches and when to release versions, balancing development efficiency and stability. Or EA can optimize multi-objective quality prediction models for different stages of the software lifecycle. In addition to software engineering, LLM and EA have the potential to promote the development of automatic circuit design and provide new innovative ideas for circuit designers. LLM can analyze a large amount of existing circuit design documents to extract key information such as functional modules and interface specifications to define the design space for EA. EA can optimize circuit topology, component selection and layout to meet multi-objective optimization of performance, power consumption and cost. LLM can also generate initial circuit design schemes for EA with reference to existing design cases. EA further optimizes the circuit topology through LLM-guided variation and selection. Among them, the design schemes and reports generated by LLM can be used as training data to improve LLM's ability in this field. Moreover, with the development of microelectronics technology, the scale of circuits continues to expand, and LLM in large-scale input processing and EA in layered optimization may play an important role. Another example unmentioned before is the intelligent agent \cite{qian2024investigate}, which can learn and evolve its behavioral strategies through interaction with the environment. In \cite{zhang2024agent}, the inference and reflection capabilities of the intelligent agent are realized through interaction with LLM. Among them, policy optimization is achieved through prompt optimization, following a simplified evolutionary process similar to ``natural selection". Future research can explore more complex evolutionary mechanisms to aid in the self-evolution of intelligent agents.}

\section{Conclusion}


\todo{In this paper, we have undertaken a comprehensive exploration of the intersection between Evolutionary Algorithms (EAs) and Large Language Models (LLMs) in the transformative era of AI. We introduced three research paradigms: LLM-enhanced EA, EA-enhanced LLM, and applications driven by the integrated synergy of LLM and EA. These paradigms exemplify the amalgamation of LLMs and EAs in various application scenarios, showcasing their collaborative strengths in tasks such as Neural Architecture Search (NAS), code generation, software engineering, and text generation. As we look forward, the collaboration between EA and LLM has garnered increasing attention, with research focusing on enhancing mutual performance and driving task-specific improvements. Currently, existing LLMs lack the direct capability to handle complex and large-scale optimization problems, but they have shown promise in generating optimization algorithms and evaluating optimization results. Future work should focus on leveraging LLMs to solve more complex optimization problems by exploring approaches such as interpreting LLM behavior and comprehensively evaluating performance on diverse issues. The exploration of enhancing LLMs with EAs is still in its early and fragmented stages. Areas beyond prompt engineering and NAS deserve further in-depth and systematic investigation. These endeavors are crucial in advancing Evolutionary Computing to effectively address the forefront challenges in the AI community. Theoretical analysis of algorithms combining LLMs and EAs from an optimization theory perspective is essential to understand properties like convergence and complexity. In practical applications, mature collaborative paradigms have emerged where LLMs and EAs work together. These paradigms leverage the iterative search framework provided by EAs and the intelligent evolutionary operators offered by LLMs to tackle various text-based optimization problems, such as code generation and prompt engineering. Generalized code generation for intricate tasks may involve modular and interactive methods. Broadening the application scope and generalization ability of LLMs in NAS through techniques such as pretraining specialized models also merits investigation. While improvements within this established paradigm or extensions to new application domains are valuable, it is of paramount importance to transcend this paradigm and explore novel collaborative mechanisms. Overall, the collaboration between EA and LLM holds significant potential for future advancements. By further refining the interaction between these fields, we can pave the way for more efficient and effective optimization strategies and drive innovation in the broader field of artificial intelligence.}

\bibliographystyle{IEEEtran}
\bibliography{llmea}

\vfill

\end{document}